%% file: main.tex
\documentclass[10pt]{article}
\usepackage[margin=1in]{geometry}
\usepackage{amsmath,amssymb,graphicx,booktabs,natbib,microtype,xcolor}
\usepackage[colorlinks=true,allcolors=blue!55!black]{hyperref}
\usepackage{caption}
\title{\bf Correcting a learned physical invariant\\improves world-model rollouts}
\author{Richard Bao\\ Independent Researcher\\ \texttt{richardbao419@gmail.com}}
\date{}
\begin{document}
\maketitle
\input{sections/abstract}
\input{sections/introduction}
\input{sections/setup}
\input{sections/recovery}
\input{sections/refusal}
\input{sections/intervention}
\input{sections/limits}
\input{sections/related}
\input{sections/conclusion}
\input{sections/appendix}
\bibliographystyle{plainnat}
\bibliography{refs}
\end{document}

%% file: sections/abstract.tex
\begin{abstract}
World models can predict video without learning dynamics that they reliably preserve. We test
whether a frozen DreamerV3 trained only on pendulum video learns a scalar that its own latent
transition treats as approximately conserved. A label-free search recovers the same energy-like
invariant across independently trained conservative models, while the same procedure finds no
comparable invariant in matched damped models. During autonomous rollouts, this quantity drifts.
Projecting the latent state back toward its initial level set reduces rollout error in all three
conservative models, whereas matched random constraints usually increase it. These results
distinguish a dynamically meaningful invariant from a merely decodable correlate and reveal a
concrete failure mode: a world model can learn a physical constraint from pixels yet violate that
constraint when it imagines forward.
\end{abstract}

%% file: sections/introduction.tex
\section{Introduction}
\label{sec:intro}

World models learn to predict future observations and can use those predictions for planning. Good
video prediction, however, does not by itself show that a model has learned the underlying dynamics
of the system it observes. That distinction matters most when we ask the model to predict far beyond
the trajectories it saw during training.

A common way to look for physical structure is to fit a probe from the hidden state to a physical
variable. Probe accuracy tells us what information a latent representation retains, but not whether
the model's transition uses that information. Our own baseline makes this problem concrete. Six
randomly initialized DreamerV3 models contain polynomial functions of their latent state that
correlate with pendulum energy at up to $0.908$. A strong correlation can therefore appear even
before learning.

\begin{figure}[t]
\centering
\includegraphics[width=\textwidth]{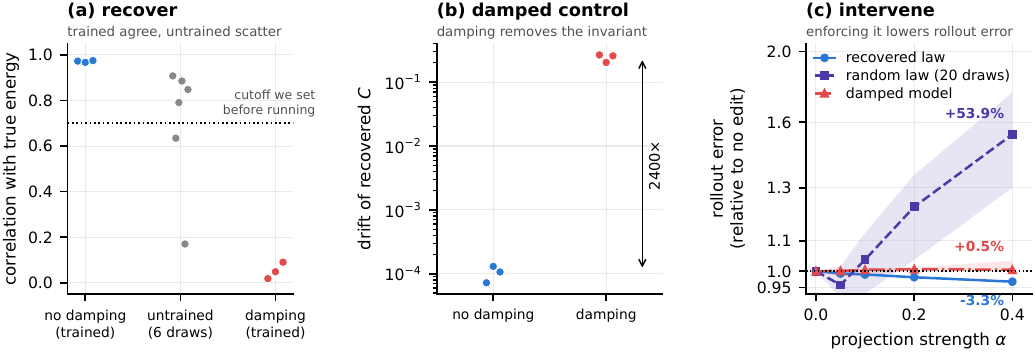}
\caption{Each point represents one model. \textbf{(a)} The three trained conservative models recover
energy correlations between $0.967$ and $0.975$, while six untrained models span $0.17$--$0.91$.
\textbf{(b)} The recovered scalar changes very little within observation-conditioned trajectories
for conservative models and much more for damped models. \textbf{(c)} Rollout error as a function of
projection strength $\alpha$, relative to no edit. Enforcing the recovered constraint lowers error;
enforcing a matched random polynomial constraint usually raises it; applying the recovered
constraint to a damped model changes little.}
\label{fig:three}
\end{figure}

We therefore ask whether a trained world model contains a scalar that its own transition
approximately preserves, and whether failures to preserve that scalar contribute to prediction
error. We study a frozen DreamerV3 \citep{hafner2023dreamerv3} trained only on $64\times64$ video of
a Gymnasium pendulum, with no physical labels and no actor or critic.

Independently trained models recover nearly identical energy-like invariants. Decodability alone
does not explain this: randomly initialized Dreamers also produce strongly energy-correlated
readouts, while matched models trained on damped dynamics contain no comparably conserved scalar.
The recovered invariant then begins to drift during autonomous imagination. Correcting that drift
improves 50-step predictions in every conservative model, whereas matched random corrections usually
make them worse. Dreamer has learned a physical constraint that its own rollout dynamics fail to
preserve.

%% file: sections/setup.tex
\section{Recovering a latent invariant}
\label{sec:setup}

\paragraph{Model and data.} We analyze the published DreamerV3 architecture
\citep{hafner2023dreamerv3}: 13.5M parameters, categorical $32\times32$ stochastic latents, a
convolutional encoder and decoder, Kullback--Leibler balancing, and unimix. We train only the world
model, with no actor or critic.

The data come from Gymnasium's \texttt{Pendulum-v1} \citep{towers2024gymnasium}. Each simulator step
renders a $500\times500$ frame, which we crop to $448\times448$ and block-average to $64\times64$. We
set the simulator state directly, sampling $\theta\sim U[-1.8,1.8]$ and
$\dot\theta\sim U[-2.2,2.2]$, and reject any trajectory that reaches the simulator's
$|\dot\theta|=8$ speed clip, since clipping would break conservation. Actions are always zero. A
single frame does not reveal velocity, so the model must infer $\dot\theta$ from the sequence.

We train on 204 trajectories of 120 frames using Adam at $10^{-4}$, batch size 16, and sequence
length 64, capped at 30 minutes of wall-clock time, which gives about 6{,}500 gradient steps. The
remaining 52 trajectories are reserved as a post-training \emph{analysis set}. Three independently
trained seeds pass checks fixed before analysis: KL divergence above 1 nat, one-step decoding at
least $4\times$ better than predicting the dataset mean, and finite rollouts.
Appendix~\ref{sec:details} gives the remaining training details. Our claims concern a DreamerV3
world model trained on pendulum video, not DreamerV3 as a full reinforcement-learning agent.

\paragraph{Three kinds of latent trajectories.} We use three latent trajectories and keep them
separate throughout.

An \emph{observation-conditioned} trajectory is the sequence of deterministic recurrent states
produced when the model receives the next real frame at every step. This is what the encoder returns
on the analysis set.

The \emph{one-step transition} $T$ is the model's autonomous recurrent update with zero action and
no new observation.

An \emph{imagination rollout} starts from one encoded state and repeatedly applies $T$ without
supplying any later frames.

We estimate the latent flow by applying $T$ once at each state of an observation-conditioned
trajectory, so the transition is autonomous even though the states come from real video.
Sections~\ref{sec:recovery} and~\ref{sec:refusal} measure conservation along observation-conditioned
trajectories. Section~\ref{sec:intervention} tests what happens to the same scalar under autonomous
imagination.

\paragraph{Latent coordinates.} After training we freeze the model. Let $h$ denote the deterministic
recurrent state on the analysis set and $\bar h$ its mean. We take the top 12 principal directions
of $h$, remove any direction with no support in the data, and let $P$ denote the resulting
orthonormal map. All extraction, projection, and perturbation operate on
\begin{equation}
z = P^\top(h-\bar h).
\label{eq:z}
\end{equation}
Let $F(z)$ be the projection through $P$ of the model's one-step displacement $T(h)-h$. Using the
model's own displacement avoids fitting a separate vector field and then analyzing that surrogate.

\paragraph{The candidate family.} We look for the invariant among degree-4 polynomials in $z$.
Writing $\varphi(z)$ for the vector of monomials up to degree 4 in the 12 latent coordinates, a
candidate is $C(z) = a^\top\varphi(z)$, and the search is over the coefficient vector $a$. There are
1819 such monomials, from $z_1$ through $z_1z_2z_3z_4$, excluding the constant term, which is
conserved trivially. The pendulum's energy is quadratic in $\dot\theta$ and needs polynomial terms
to approximate $\cos\theta$, and the latent is an unknown nonlinear encoding of
$(\theta, \dot\theta)$ rather than those coordinates themselves, so the family must be rich enough
to express energy through that distortion while staying small enough to fit.

\paragraph{The invariance criterion.} An invariant is constant along any one trajectory and differs
between trajectories. We score a candidate by
\begin{equation}
\text{ratio}(C) \;=\;
\frac{\text{mean within-trajectory variance of } C}{\text{total variance of } C},
\label{eq:ratio}
\end{equation}
which is $0$ for a perfectly conserved scalar and near $1$ for one that wanders as much within a
trajectory as across the dataset. The denominator keeps the criterion non-trivial: without it,
$C = 0$ would score perfectly while distinguishing nothing.

Both variances are quadratic forms in $a$. With $W$ the mean within-trajectory covariance of the
features and $T$ their total covariance, Equation~\ref{eq:ratio} is $a^\top W a / a^\top T a$, a
Rayleigh quotient. Its minimiser is the eigenvector of the generalized problem
$W a = \lambda\, T a$ with the smallest eigenvalue, and $\lambda$ is the invariance ratio itself. One
eigendecomposition returns the whole family, ranked from most to least conserved.

\paragraph{Selecting among conserved candidates.} Conservation alone does not pick out a unique
scalar. If $C$ is conserved then so is any function of it, so $E$, $E^2$ and mixtures of $E$ with
other conserved quantities all sit near the top of the ranking, and the leading eigenvector is
generally some combination of them. We therefore use flow alignment as a secondary criterion among
the leading candidates, fitting $C$ jointly with an antisymmetric operator $B$ such that
\begin{equation}
F \approx B\nabla C, \qquad B^\top = -B.
\label{eq:pair}
\end{equation}
This mirrors the Hamiltonian relation between a conserved quantity and the flow it generates: an
antisymmetric operator maps the gradient of $C$ to a direction tangent to its level set
\citep{arnold1989mathematical}. Because $\nabla C^\top B \nabla C = 0$ for any antisymmetric $B$, a
scalar satisfying Equation~\ref{eq:pair} is constant along the flow by construction. The criterion
also separates candidates the invariance ratio cannot: $E^2$ is exactly as conserved as $E$ but does
not pair with the same $B$.

The fit is bilinear, so fixing either $a$ or $B$ makes the other a least-squares problem and we
alternate. It searches within the top eight eigenvectors from Equation~\ref{eq:ratio}, so the
recovered $C$ has eight effective degrees of freedom rather than 1819. That matters for a search run
on 52 trajectories, where a free fit over the full basis could drive the in-sample ratio to zero by
overfitting. Appendix~\ref{sec:details} isolates the contribution of the flow criterion:
conservation drives most of the recovery and intervention effect, while flow alignment mainly
reduces variation across seeds.

\paragraph{Reference energy.} We compare $C$ with the textbook pendulum energy
$E = \tfrac{1}{6}\dot\theta^2 + 5\cos\theta$. Gymnasium uses a semi-implicit integrator, so it
conserves a nearby shadow Hamiltonian $\tilde H = H + O(\Delta t)$ rather than $E$ itself
\citep{hairer2006geometric}. In our trajectories this produces about 12\% relative oscillation in
$E$ without secular drift, which we treat as a noise floor.

The search sees only latent states and the model's own one-step transition. It never sees $\theta$,
$\dot\theta$ or $E$, and nothing in Equations~\ref{eq:ratio} and~\ref{eq:pair} refers to them. Ground
truth enters afterward, when we score the recovered $C$ against $E$. We selected the extraction
dimension on three development models and fixed it before evaluating the three models reported here.

\paragraph{Code and data.} The extraction code, the run logs behind every figure and table, and
the source of this paper are at \url{https://github.com/Zarand3r/world-model-invariants}. The figures
regenerate from the committed logs without a GPU.

\paragraph{Preliminary experiments.} We developed the extraction procedure in preliminary
experiments on smaller recurrent models. Those experiments motivated the untrained, dissipative, and
intervention controls used here, and we do not use them as evidence for the DreamerV3 results.

%% file: sections/recovery.tex
\section{Training makes invariant recovery reproducible}
\label{sec:recovery}

At the extraction dimension fixed in advance, the three trained conservative models recover scalars
with $|\rho|_E = 0.973$, $0.967$, and $0.975$. Along observation-conditioned trajectories the
recovered $C$ has within-trajectory variance equal to $1.1\times10^{-4}$ of its total variance, so it
is nearly constant within a trajectory while still varying across trajectories.
Table~\ref{tab:refusal} reports this quantity as the drift of $C$. The held-out invariance ratio in
the next row evaluates the invariance-only candidate rather than the jointly fitted $C$, so the two
rows measure different objects.

\begin{table}[t]
\centering\small
\caption{The same extraction pipeline applied to conservative and damped models. Each range covers
three seeds. No conservative seed overlaps any damped seed on any row.}
\label{tab:refusal}
\begin{tabular}{lccc}
\toprule
statistic & conservative & damped & median ratio \\
\midrule
$|\rho|_E$ & 0.967--0.975 & 0.018--0.090 & $20\times$ \\
drift of $C$ & $0.7$--$1.3\times10^{-4}$ & 0.203--0.266 & $2414\times$ \\
held-out invariance ratio & 0.0027--0.0121 & 0.983--0.994 & $282\times$ \\
pairing residual & 0.829--0.865 & 0.934--0.957 & n/a \\
\bottomrule
\end{tabular}
\end{table}

The same pipeline applied to six randomly initialized DreamerV3 models gives $|\rho|_E$ from $0.170$
to $0.908$. Four of the six exceed the $0.7$ threshold we had inherited from the earlier
architecture, so that threshold does not identify learning in DreamerV3. On a one-degree-of-freedom
conservative pendulum, energy is essentially the only nontrivial scalar that is constant within a
trajectory and differs across trajectories, so any latent that retains $(\theta,\dot\theta)$ can
support a strong energy-correlated readout \citep{hewitt2019control,belinkov2022probing}.

The trained and untrained groups do separate on two statistics without overlap. Every trained model
scores above every untrained draw in energy correlation ($0.967$ versus $0.908$ at the closest
pair), and every trained model has a lower pairing residual than every untrained draw ($0.865$
versus $0.912$). With three trained seeds and six random draws these non-overlaps are suggestive
rather than statistically decisive, and we attach no $p$-value to them.

The more robust difference is reproducibility. The three trained models cluster between $0.967$ and
$0.975$, while the six untrained models span $0.170$--$0.908$. Training therefore appears to make
the recovered scalar reproducible across independently trained models rather than merely making
energy decodable.

%% file: sections/refusal.tex
\section{The invariant disappears under matched damping}
\label{sec:refusal}

A search that produces a conserved-looking scalar for almost any latent representation would explain
Section~\ref{sec:recovery} without telling us anything about the learned dynamics. Matched models
trained on a damped pendulum test that possibility.

For the damped system every trajectory converges to the same fixed point. Any continuous scalar that
stays constant along all trajectories in the basin must be constant throughout that basin, so the
state has no nontrivial first integral for the search to recover.

\paragraph{Choosing the damping level.} Very strong damping creates a different problem: trajectories
collapse into a small neighborhood of the fixed point, the latent becomes rank-deficient, and
directions appear constant for trivial reasons. We fixed the damping level using only simulator
states, before training any damped model, taking the largest damping ratio whose late-window spread
across trajectories remained above 25\% of the conservative reference. A stronger value,
$\zeta=0.15$, leaves 0\% of that spread and fails the criterion. The rule selects $\zeta=0.03$,
which removes 17\% of the energy per period while retaining 34\% of the conservative spread. All
three damped models pass the same training checks as the conservative models.

\paragraph{Four statistics separate without overlap.} Table~\ref{tab:refusal} shows the result. The
recovered $C$ varies about $2400\times$ more within damped trajectories than within conservative
ones, and the damped models have a held-out invariance ratio of $0.983$--$0.994$ against
$0.0027$--$0.0121$ for the conservative models. Within the degree-4 family we search, the damped
models contain no approximately conserved candidate.

The two arms share architecture, training budget, observation modality, and extraction procedure.
The dynamics in the training data are the only difference. This does not prove that a damped model
contains no conserved quantity of any possible form; it shows that the pipeline does not manufacture
one when the underlying dynamics remove the corresponding conservation law.

%% file: sections/intervention.tex
\section{Correcting invariant drift improves rollouts}
\label{sec:intervention}

If invariant drift contributes to rollout error, correcting that drift should improve prediction. If
the recovered scalar is only a descriptive correlate, enforcing it has no reason to help. The
recovered $C$ stays nearly constant along observation-conditioned trajectories and drifts once the
model rolls forward on its own, so the test is available.

At every imagined step we project the latent state back toward the level set defined by its initial
value:
\begin{equation}
z \leftarrow z - \alpha\bigl(C(z)-C_0\bigr)\frac{\nabla C(z)}{\|\nabla C(z)\|^2},
\label{eq:edit}
\end{equation}
where $C_0$ is the value at the first encoded state. We map the correction back through $P$, leaving
the part of $h$ outside the extracted subspace unchanged. The correction runs inside the imagination
loop, does not touch the decoder, and the model is never retrained.

\paragraph{Scoring and controls.} We sweep $\alpha \in \{0, 0.05, 0.1, 0.2, 0.4\}$, identical for
every arm, and score the slope of rollout error across the whole grid; taking the best $\alpha$
would give each arm five chances to look good. Rollout error is pixel mean squared error against the
analysis set over 50 imagined steps. We fit $C$ and evaluate the correction on the same analysis
trajectories, so the absolute effect is in-sample with respect to invariant fitting; the comparison
with random constraints remains matched, since both use the same trajectories.

We compare three interventions: each conservative model's own recovered $C$, a random polynomial
from the same degree-4 basis matched in norm, and each damped model's own recovered $C$. We draw 20
random polynomials and apply the same 20 to each of the three conservative models, giving 60
model--constraint evaluations from 20 distinct constraints.

Figure~\ref{fig:three}c shows the dose response. At the strongest projection the recovered
constraint lowers rollout error by 2.9\%, 3.3\%, and 3.5\% across the three conservative models.
Across the 60 random-constraint evaluations the median change is $+53.9\%$, and 7 of 60 lower the
error. The same edit on the damped models changes error by $+0.5\%$. The recovered arm is also far
more consistent across seeds: its per-model slopes span $0.016$ against $3.71$ for the random
evaluations.

\begin{figure}[t]
\centering
\includegraphics[width=0.52\textwidth]{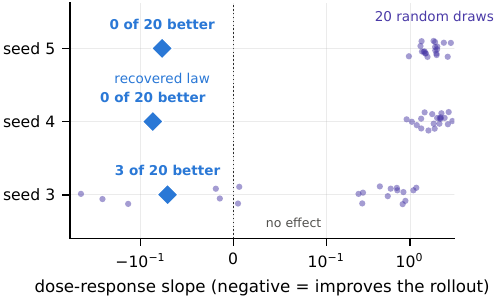}
\caption{Each row compares one model with its own null distribution. The 20 small marks are
norm-matched random degree-4 constraints, using the same draws across models; the diamond is that
model's recovered constraint. Labels count the random constraints with a better dose-response slope
than the recovered one.}
\label{fig:null}
\end{figure}

\paragraph{Specificity is not uniform.} Figure~\ref{fig:null} compares each model with its own
20-draw null. On two models no random constraint achieves a better dose-response slope than the
recovered one. On the third, 3 of 20 do, 7 of 20 lower the error at the strongest projection, and
the best random draw reaches $-21.9\%$ against $-2.9\%$ for the recovered constraint. We do not know
why this model differs.

Enforcing the recovered invariant improves all three conservative models by 2.9--3.5\%, while
matched random constraints usually hurt. The improvement is specific to the recovered invariant on
two models and not on the third.

The correction is not simply concentrated in the highest-variance latent directions. At a 40-step
horizon, variance rank is negatively correlated with causal sensitivity in all three models
($-0.364$, $-0.259$, $-0.406$), while the correction preferentially acts along higher-sensitivity
directions ($+0.385$, $+0.692$, $+0.287$). Appendix~\ref{sec:sensitivity} gives the full perturbation
and subspace analyses.

%% file: sections/limits.tex
\section{Limitations}
\label{sec:limits}

This study covers one physical system, one degree of freedom, one architecture, and a much shorter
training schedule than a standard DreamerV3 run. Three trained seeds and six random draws are enough
to show non-overlapping ranges but not enough for a precise effect size, and we report ranges rather
than confidence intervals throughout.

The intervention effect is modest and not uniform. Enforcing the recovered invariant improves all
three conservative models, but relative to matched random constraints the improvement is specific on
only two of them. We fit $C$ and evaluate the correction on the same analysis trajectories, so the
absolute effect is in-sample with respect to invariant fitting.

Conservation, not the full extraction rule, carries the result. Selecting $C$ by the invariance
criterion alone recovers energy nearly as well and intervenes slightly better; flow alignment mainly
reduces variation across seeds (Appendix~\ref{sec:details}). The pairing residual does not track
recovery quality across extraction dimension and should be read as a diagnostic rather than a
confidence score.

The pendulum trajectories cross the separatrix, so 17\% rotate rather than librate. We have not
tested contacts, multiple interacting objects, nonzero actions, or systems whose frequency varies
enough to identify a frequency-weighted accumulation of invariant error; the present experiments do
not distinguish weighted from unweighted accumulation (Appendix~\ref{sec:sweeps}).

%% file: sections/related.tex
\section{Related work}
\label{sec:related}

Probes establish that a variable can be recovered from a representation, but not that the model's
dynamics use or preserve it \citep{alain2017probing}. Control tasks \citep{hewitt2019control} and
later critiques \citep{belinkov2022probing} show that probe accuracy mixes what the representation
stores with what the probe can learn, and our untrained baseline is the same problem in a different
form: four of six randomly initialized models exceed the threshold we had previously used. Work on
emergent world representations goes further by intervening on the representation
\citep{li2023othello,nanda2023othello}, which is the logic our rollout correction follows.

A separate literature imposes physical structure by design. Equation-discovery methods such as SINDy
\citep{brunton2016sindy} infer governing equations from observed trajectories, and Hamiltonian and
Lagrangian networks \citep{greydanus2019hnn,cranmer2020lnn} build conservation into an architecture
so that it holds by construction. We instead ask whether such structure emerges in a conventionally
trained world model, and analyze it after the fact using standard linear-algebraic tools related to
Koopman and dynamic mode methods \citep{schmid2010dmd}.

World models are normally evaluated through prediction quality or control performance
\citep{hafner2023dreamerv3,ha2018world,schrittwieser2020muzero}. Those metrics do not reveal whether
the latent transition preserves a physical constraint it has learned. Our intervention tests whether
violating a recovered latent constraint itself contributes to rollout error, and the damped arm
tests whether the extraction procedure returns a conserved-looking scalar when the training dynamics
have no such invariant. The disentanglement literature \citep{locatello2019challenging} documents how
readily an appealing structural claim about representations survives without that kind of control.

%% file: sections/conclusion.tex
\section{Conclusion}

A DreamerV3 trained only on pendulum video learns an energy-like latent quantity that its own
dynamics approximately preserve on observation-conditioned trajectories. Correlation alone does not
establish this: randomly initialized models also admit strongly energy-correlated readouts. The
stronger evidence comes from matched damping and intervention. The invariant disappears when the
training dynamics are dissipative, and correcting its drift during autonomous imagination improves
prediction in every conservative model.

The result exposes a specific gap between learning physical structure and respecting it during
rollout. A world model can encode a useful dynamical constraint yet violate that constraint when it
predicts autonomously. Whether this phenomenon survives in systems with actions, contacts, and
multiple interacting objects is the main open question.

%% file: sections/appendix.tex
\appendix

\section{Experimental details and extraction ablations}
\label{sec:details}

\paragraph{Training.} We set free bits to 0 rather than DreamerV3's reference default of 1.0. The
measured KL divergence settles near 1.8 nats, well above collapse, so the floor is inactive here. All
three trained models pass the acceptance checks; none were discarded.

\paragraph{Data split.} Training uses trajectories 0--203 of the 256-trajectory dataset. The
remaining 52 form the analysis set. We fit $C$ on those 52 trajectories and evaluate the
Section~\ref{sec:intervention} correction on the same 52, so the absolute effect is in-sample with
respect to invariant fitting. The recovered and random-constraint arms use the same trajectories.

\paragraph{Scoring the correction over a grid.} We score the slope of rollout error over the fixed
$\alpha$ grid rather than the minimum on that grid. Taking the minimum selects the most favourable of
five points after seeing the curve. In one diagnostic run that rule would have reported a 5.9\%
improvement for an arm whose error rose monotonically to twice baseline.

\paragraph{Flow alignment contributes reproducibility.} Holding the candidate family, data,
dimension, degree, and $\alpha$ grid fixed, we select $C$ either by the invariance criterion alone or
by the joint criterion. Conservation alone recovers energy at $0.950$ against $0.973$ for the joint
rule, and gives a slightly better mean intervention slope ($-0.087$ against $-0.077$, winning on two
of three models). The difference is in spread: recovery varies by $0.035$ across seeds under
conservation alone and by $0.008$ under the joint rule.

\paragraph{The pairing residual does not track recovery.} As the extraction dimension rises from 6 to
12, energy recovery improves from $0.721$ to $0.967$ while the pairing residual worsens from $0.738$
to $0.876$. Recovery already reaches $0.967$ at dimension 8, where the residual is lower at $0.813$.
The added dimensions do not carry energy while resisting the Hamiltonian fit: directions 7--12
account for 44.9\% of the residual and 43.5\% of the flow. Figure~\ref{fig:ldsweep} gives the sweep.

\section{Latent sensitivity analyses}
\label{sec:sensitivity}

For each direction $u$ of the extracted subspace we measure its variance and the damage a
displacement along it does to an imagination rollout,
\begin{equation}
V(u) = \operatorname{Var}(u^\top z), \qquad
D_H(u) = \mathbb{E}\bigl[L_H(z+\epsilon u) - L_H(z)\bigr],
\label{eq:leverage}
\end{equation}
with $L_H$ the $H$-step imagination error against the analysis set.

\paragraph{Calibrating the displacement.} At $\epsilon = 0.05|z|$ the damage is about $-0.5\%$ of
baseline with the sign varying across directions. Imagination is deterministic, so repeated
unperturbed rollouts are identical and these values are not sampling noise. Damage for the leading
direction runs $-0.5\%$, $+3.1\%$, $+18.1\%$, $+46.4\%$, and $+128.8\%$ at $\epsilon/|z|$ of $0.05$,
$0.10$, $0.25$, $0.50$, and $1.00$. We use $\epsilon = 0.25|z|$, the smallest displacement whose
damage exceeds 10\% of baseline while the response still grows smoothly.

A displacement does not decay under imagination. After 40 steps it retains $6.3\times$ its initial
size at $\epsilon = 0.05|z|$, $2.8\times$ at $0.25|z|$, and $1.9\times$ at $0.5|z|$, as medians over
the three models.

\begin{figure}[h]
\centering
\includegraphics[width=\textwidth]{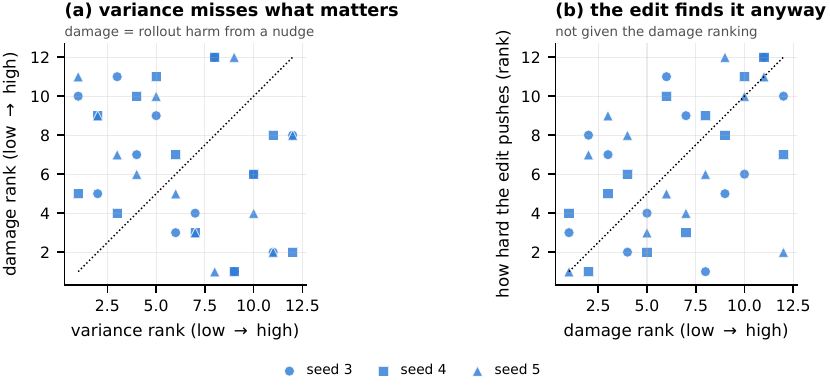}
\caption{Each point is one of the 12 extracted directions, ranked within its own model; the dotted
line marks equal ranks. \textbf{(a)} Variance rank against damage rank at a 40-step horizon.
\textbf{(b)} Damage rank against the magnitude of the invariant correction along that direction.}
\label{fig:leverage}
\end{figure}

\paragraph{Variance and sensitivity at different horizons.} At a 40-step horizon the two rankings
anticorrelate on all three models ($-0.364$, $-0.259$, $-0.406$), with damage spanning
$13.5\times$ to $118.8\times$ between the most and least consequential direction
(Figure~\ref{fig:leverage}a). The ranking replicates: split-half rank correlation of $D_H$ has median
$0.95$ across the settings we tried, with a minimum of $0.52$ at the smallest displacement and
longest horizon. The disagreement with variance does not extend to long horizons. Pooling three
models and three displacements, median $\rho(V, D_H)$ is $-0.252$ at horizon 20, negative in 9 of 9
settings; $-0.196$ at horizon 50, negative in 5 of 9; and $-0.028$ at horizon 100, negative in 5 of
9, with one setting reaching $+0.727$. Once a rollout has diverged, high-amplitude directions
dominate the error and the rankings realign.

The correction pushes along higher-damage directions at $+0.385$, $+0.692$, and $+0.287$
(Figure~\ref{fig:leverage}b), without having been given that ranking. The correlations stay well
below $1$, so the correction is only partly concentrated there. All of this is measured on the
conservative models.

\paragraph{Basis-independent properties reproduce better than coordinate selectors.} The three models
agree closely on quantities that need no choice of basis: energy recovery varies by $0.008$,
intervention slopes by $0.016$, and the top-3 eigenvalue mass of
$M_C = \mathbb{E}[gg^\top/\|g\|^2]$ with $g = \nabla C$ by $0.046$ ($0.448$, $0.441$, $0.487$).
Selectors on individual coordinates are less stable: ranking coordinates by $\nabla C$ and keeping
the top three gives intervention slopes of $-0.075$, $+0.016$, and $-0.089$, beating a variance
ranking on two models and changing sign across seeds. An orthogonal projector onto the top eigenspace
of $M_C$ beats the variance selector on all three models, though on one it sits at the 41st
percentile of 100 random rank-3 subspaces. Since the correction is rank one at each state, the
spectrum of $M_C$ measures how much $\nabla C$ rotates as the state changes, and the amount is
similar across models. This is consistent with the models encoding the same scalar in different
latent coordinate systems, although we do not directly align their representations.

\section{Additional sweeps}
\label{sec:sweeps}

\begin{figure}[h]
\centering
\begin{minipage}[t]{0.36\textwidth}
\centering
\includegraphics[width=\textwidth]{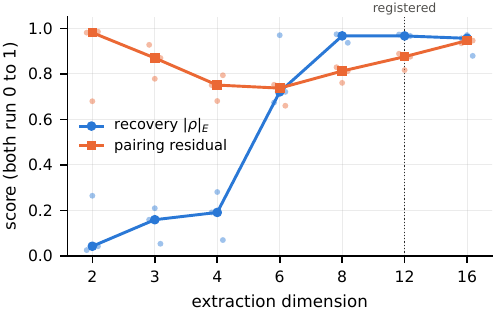}
\caption{Energy recovery and pairing residual against extraction dimension. Lines are medians over
the three models; points are individual models. Both quantities are dimensionless.}
\label{fig:ldsweep}
\end{minipage}\hfill
\begin{minipage}[t]{0.58\textwidth}
\centering
\includegraphics[width=\textwidth]{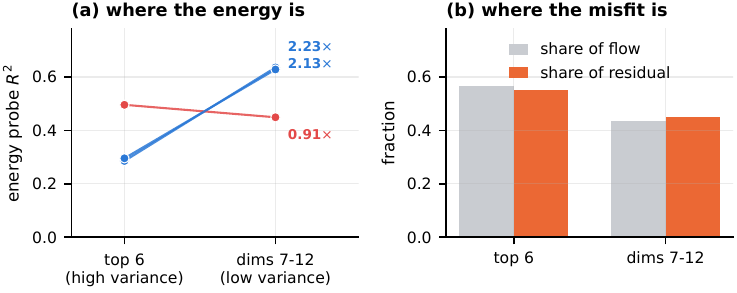}
\caption{\textbf{(a)} Energy-probe $R^2$ for the top six principal directions against directions
7--12, per model. \textbf{(b)} Fraction of latent flow and pairing residual carried by each group,
medians over three models.}
\label{fig:lowvar}
\end{minipage}
\end{figure}

\paragraph{Energy outside the highest-variance directions.} On two of three models, principal
directions 7--12 carry about twice the energy information of the top six ($2.23\times$ and
$2.13\times$); on the third the ratio reverses to $0.91$. We report the models separately because a
median would hide the sign change. Figure~\ref{fig:lowvar} shows the decomposition.

\paragraph{Accumulated invariant error.} Accumulated invariant error predicts which rollouts degrade
most: an integral predictor reaches $R^2 = 0.187$ against $0.021$ for a fitted power law in time at
the horizon where both perform best, and shuffling destroys the relation. The current experiments do
not distinguish weighted from unweighted accumulation. In a nested comparison with 400 bootstrap
resamples, the weighted term adds explanatory power beyond the unweighted term in 1 of 9
seed--horizon combinations, and the unweighted beyond the weighted in 0 of 9, at $n = 512$.

